\documentclass[11pt]{article}

\usepackage[preprint]{acl}

\usepackage{microtype}
\usepackage{amsmath}
\usepackage{amssymb}
\usepackage{graphicx}
\usepackage[table,xcdraw]{xcolor}
\usepackage{booktabs}
\usepackage{fontspec}
\usepackage{tikz}
\usetikzlibrary{shapes.geometric, arrows}
\usepackage[english,bidi=default]{babel}

\def\bng{\bngx}

\font\bngx=bang10

\def\*#1*#2{o\null{#2}{#1}}

\def\sh#1{\setbox0=\hbox{#1}%
     \kern-.02em\copy0\kern-\wd0
     \kern.04em\copy0\kern-\wd0
     \kern-.02em\raise.0433em\box0 }

\title{PyBangla at BLP-2025 Task 2: Enhancing Bangla-to-Python Code Generation with Iterative Self-Correction and Multilingual Agents}

\author{
  \textbf{Jahidul Islam}, 
  \textbf{Md Ataullha}, 
  \textbf{Saiful Azad} \\
  Department of Computer Science and Engineering \\
  Green University of Bangladesh, Dhaka, Bangladesh \\
  \texttt{jahidul.cse.gub@gmail.com} \quad
  \texttt{ataullha00@gmail.com} \quad
  \texttt{saiful@cse.green.edu.bd}
}

\begin{document}

\maketitle

\begin{abstract}
LLMs excel at code generation from English prompts, but this progress has not extended to low-resource languages. This paper addresses the challenge of Bangla-to-Python code generation by introducing BanglaCodeAct, an agent-based framework that leverages multi-agent prompting and iterative self-correction. Unlike prior approaches that rely on task-specific fine-tuning, BanglaCodeAct employs an open-source multilingual LLM within a Thought–Code–Observation loop, enabling the system to dynamically generate, test, and refine code from Bangla instructions.
We benchmark several prominent small-parameter open-source LLMs and evaluate their effectiveness on the mHumanEval dataset for Bangla NL2Code. Our results show that Qwen3-8B, when deployed with BanglaCodeAct, achieves the best performance, with a pass@1 accuracy of 94.0\% on the development set and 71.6\% on the blind test set. These findings establish a new benchmark for Bangla-to-Python translation and highlight the potential of agent-based reasoning for reliable code generation in low-resource languages.. Experimental scripts made publicly available at \href{https://github.com/jahidulzaid/PyBanglaCodeActAgent}{github.com/jahidulzaid/PyBanglaCodeActAgent}
\end{abstract}

\section{Introduction}
\label{sec:introduction}

\textbf{Large Language Models (LLMs)} has created a paradigm shift in software engineering, automating complex coding tasks, and democratizing programming for a larger audience \cite{chen2021evaluating}. Natural Language-to-Code (NL-to-Code) generation \cite{yin2022natural}, once a distant goal, is now a tangible reality, with systems capable of producing functional code from simple English descriptions.
The vast majority of these advances remain linguistically monolithic, centered almost exclusively on English, leaving other languages behind.. This linguistic bias creates a significant accessibility gap for millions of learners worldwide whose primary language is not English. For speakers of low-resource languages like \textbf{Bangla}, the seventh most spoken language globally.

To address this critical issue, we introduce the \textbf{BanglaCodeAct Agent}, a ReAct agent framework designed for cross-lingual code generation from Bangla instructions into executable Python. Instead of relying on task-specific fine-tuning, our approach leverages the emergent multilingual reasoning capabilities of a general-purpose open-source LLMs within an iterative, self-correcting loop.

We address 3 research questions:

\normalsize{
\begin{enumerate}
    \item \textbf{RQ1:} How can an agent-based framework be designed to generate Python code from natural language instructions in Bangla?
    \item \textbf{RQ2:} Can a general-purpose, multilingual LLMs be effectively prompted to perform cross-lingual code generation in a zero-shot setting, without the need for task-specific datasets?
    \item \textbf{RQ3:} How does incorporating an iterative \textit{Thought-Code-Observation} loop within a robust execution environment affect the reliability and correctness of the generated code?
\end{enumerate}}

\normalsize
\section{Related Work}

This research is positioned at the confluence of several rapidly advancing domains: automated code generation, the development of specialized large language models for programming, and the specific challenges within Natural Language Processing (NLP) for low-resource languages like Bangla. Our work synthesizes insights from these areas to address a novel problem: agent-driven, cross-lingual code generation from a low-resource language.

The introduction of the Transformer architecture \citep{vaswani2017attention} created major progress in this field. This led to the development of Large Language Models (LLMs) trained on vast web-scale corpora. Models like OpenAI's Codex, the engine behind GitHub Copilot \citep{chen2021evaluating}, and DeepMind's AlphaCode \citep{li2022alphacode}, which achieved competitive performance in programming contests. However, a significant limitation of this era has been a reliance on English-centric data and evaluation benchmarks.

Building on the success of general-purpose LLMs, a new wave of models has been specifically trained or fine-tuned for programming tasks. Notable examples include CodeLlama \citep{roziere2023code}; StarCoder \citep{li2023starcoder}; and DeepSeek Coder \citep{guo2024deepseek}.

To reduce hallucination and improve factual grounding, Retrieval-Augmented Generation (RAG) retrieves relevant documents from an external knowledge base and supplies them as context to the LLMs\cite{lewisretrieval}. Corrective RAG (CRAG) introduces a lightweight retrieval evaluator to augment retrieved documents, improving the correctness of the generation process \citep{yan2024corrective}.


Bangla NLP faces persistent gaps that make NL2Code especially challenging. 
First, \textbf{data scarcity}: large-scale parallel corpora of Bangla programming instructions and code are virtually absent \citep{zhong2024opportunities, raihan-etal-2025-mhumaneval}. 
Second, \textbf{morphological complexity}: Bangla’s rich inflectional system makes natural language instructions harder to parse into precise logical forms compared to English \citep{bhattacharjee2023banglabert}. 
Prior LLM-based approaches, trained or fine-tuned primarily on English or multilingual data, often fail to capture these nuances, resulting in low accuracy and unstable performance in Bangla NL2Code tasks \citep{chen2021evaluating,li2022alphacode}.  

Our work addresses these gaps by introducing \textbf{BanglaCodeAct}, which directly leverages the multilingual reasoning abilities of general-purpose LLMs in a self-correcting loop, without requiring costly Bangla-specific fine-tuning or large annotated datasets.


\section{Dataset and Evaluation Metrics}
\label{sec:Dataset}
The task involves translating Bangla natural language programming instructions into Python code, ensuring functional correctness by passing associated test cases. This setup mirrors typical NL2Code challenges but places a specific emphasis on low-resource language understanding and algorithmic reasoning in Bangla \cite{raihan-etal-2025-blp}. To evaluate this translation process, we employ the \textbf{mHumanEval dataset} \cite{raihan-etal-2025-mhumaneval}, which is tailored for Bangla-to-Python code generation. The dataset consists of natural language programming problems in Bangla, each paired with a corresponding Python implementation and unit test cases that serve as an objective correctness signal. It covers a wide range of fundamental programming concepts, including algorithmic reasoning, control structures, data manipulation, and function design. The sample structure of the dataset is presented in Table~\ref{tab:dataset_sample}.

\subsection{Evaluation Metric}
The primary metric for our evaluation is \textbf{pass@1} on the HumanEval benchmark \cite{raihan-etal-2025-mhumaneval}. A generated code snippet is considered a ``pass" if it executes without error and satisfies all provided assertions in the `test\_list` for that problem. 

\[
\text{pass@}k := \mathbb{E}_{\text{problems}} \left[ 1 - \frac{\binom{n - c}{k}}{\binom{n}{k}} \right]
\]

\begin{table*}[ht]
    \centering
    \scriptsize
    \begin{tabular}{@{}l p{0.35\textwidth} p{0.30\textwidth}@{}}
        \toprule
        \textbf{ID} & \textbf{Instruction (Bengali)} & \textbf{Test Cases} \\
        \midrule
        1 & {\bng EkiT phaNNGshn ilkhun Ja priikKa kreb pRdt/t is/TRNNG pYailneDRam ikna. khail is/TRNNGk pYailneDRam iHeseb gNY Heb.} \newline Example: \texttt{is\_palindrome(s)} & 
        \begin{tabular}[t]{@{}l@{}}
            assert is\_palindrome(``TENET") == True \\
            assert is\_palindrome(``Bangla") == False \\
            assert is\_palindrome(`` ") == True
        \end{tabular} \\
        \midrule
        2 & {\bng EkiT phaNNGshn ilkhun Ja EkiT is/TRNNG-Er medhY thaka shb/dguelaek Uel/Ta ker sajaeb.} \newline Example: \texttt{reverse\_words(string)} &  
        \begin{tabular}[t]{@{}l@{}}
            assert reverse\_words(``hello")==``hello" \\
            assert reverse\_words(`` a b ") == ``b a" \\
            assert reverse\_words(``hello world") ==  \newline ``world hello"
        \end{tabular} \\
        \midrule
        3 & {\bng EkiT pa{I}thn phaNNGshn ilkhun Ja idJe du{I}iT puur/NsNNGkhYar ibpriit icn/H Aaech ikna ta priikKa kra JaJ.}\newline Example: opposite\_Signs(n1, n2) &  
        \begin{tabular}[t]{@{}l@{}}
            assert opposite\_Signs(1,-2) == True \\
            assert opposite\_Signs(3,2) == False \\
            assert opposite\_Signs(-10,-10) == False
        \end{tabular} \\
        \bottomrule
    \end{tabular}
    \caption{The dataset for Shared Task 2 (Code Generation) includes Bengali programming instructions, the corresponding Python code implementations, and test cases designed for validation.}
    \label{tab:dataset_sample}
\end{table*}

\section{Methodology}
\label{sec:methodology}

In this work we introduce an agent framework, \textbf{BanglaCodeAct Agent}, for cross-lingual code generation. The primary objective is to translate natural language programming instructions articulated in a low-resource language, \textbf{Bangla (Bengali)}, into executable Python code. The methodology hinges on a powerful multilingual Large Language Models (LLMs) integrated into an iterative reasoning and self-correction loop, enabling it to bridge the semantic gap between Bangla prose and Python's formal syntax.

\subsection{Models and Baselines}
\label{appendix:Baselines}

We compare the performance of our proposed \textbf{BanglaCodeAct Agent} against several baselines to evaluate the contribution of its components and to situate its performance relative to other approaches. First, \textbf{Zero-Shot Prompting} serves as a direct baseline where the model is given only the system prompt and the user task (Bangla instruction plus test cases) and is asked to generate the solution in a single turn. This approach achieves varying results across models: Qwen/Qwen3-8B obtains 36\%, Qwen-Coder-7B reaches 51\%, TigerLLM-1B-it \cite{raihan2025tigercoder} it achieves 11\%, and Llama-3.1-8B performs the best at 77\%. Next, \textbf{Few-Shot Prompting} provides the model with a small number of solved examples in the prompt to help it generalize to new problems. Performance here also varies, with Qwen3-8B achieving 46\%, Qwen2.5-Coder-7B reaching 51\%, and Llama-3.1-8B again performing strongly at 77\%. DeepSeek-Coder-V2-Lite shows competitive results with a pass@1 of 73.0\%. The \textbf{Self-Consistency} method leverages Qwen/Qwen3-8B to generate multiple independent solutions for the same problem and selects the final answer through majority voting, without using any iterative feedback loop. Finally, our full proposed framework, the \textbf{BanglaCodeAct Agent}, based on Qwen/Qwen3-8B, significantly outperforms these baselines with a 94\% success rate. This agent employs an iterative \textit{Thought-Code-Observation} loop, allowing it to self-correct based on execution feedback until all test cases are satisfied.

\begin{table}[ht]
\centering
\small
\begin{tabular}{>{\columncolor[HTML]{D9EAD3}}l l}
\toprule
\rowcolor[HTML]{CFE2F3} 
\textbf{Parameters}        & \textbf{Value}   \\ \midrule
\rowcolor[HTML]{F9F9F9} 
Max tokens                 & 8192             \\
Temperature                & 0.7              \\
Top-p                      & 0.9              \\
Best-of                    & 1                \\
Repetition penalty         & 1.05 (CoT)       \\
Decoding                   & Self-consistency ($n=5$)  \\
Num paths                  & 16 / 5 (SC)      \\
Seed                       & 42               \\
Timeout                    & 5 Seconds        \\
Retries                    & 25               \\ \bottomrule
\end{tabular}
\caption{Inference hyperparameters. These decoding and sampling parameters control output length, diversity, reproducibility, and error handling.}
\label{tab:hyper}
\end{table}

The experiments were executed with inference controlled by the hyperparameters presented in Table~\ref{tab:hyper}.

The agent’s core is the \textbf{Qwen/Qwen3-8B} model, a multilingual LLM capable of zero-shot Bangla-to-logic translation and reasoning. To enable efficient multi-turn reasoning, we deploy it with the \textbf{vLLM inference engine}, leveraging \textbf{tensor parallelism} and \textbf{prefix caching} for reduced latency and high throughput \cite{kwon2023efficient}.

\subsection{Cross-Lingual BanglaCodeAct Agent Framework}
We employ the Code Acting (\texttt{CodeAct}) paradigm to structure the agent's problem-solving process. This approach transforms code generation from a single-shot task into a dynamic, multi-step dialogue between the agent and a code interpreter. The agent operates on a \textit{Thought-Code-Observation} cycle (as illustrated in Fig. \ref{fig:codeact-methodology}):

\begin{figure}[ht]
    \centering
    \includegraphics[width=0.70\linewidth]{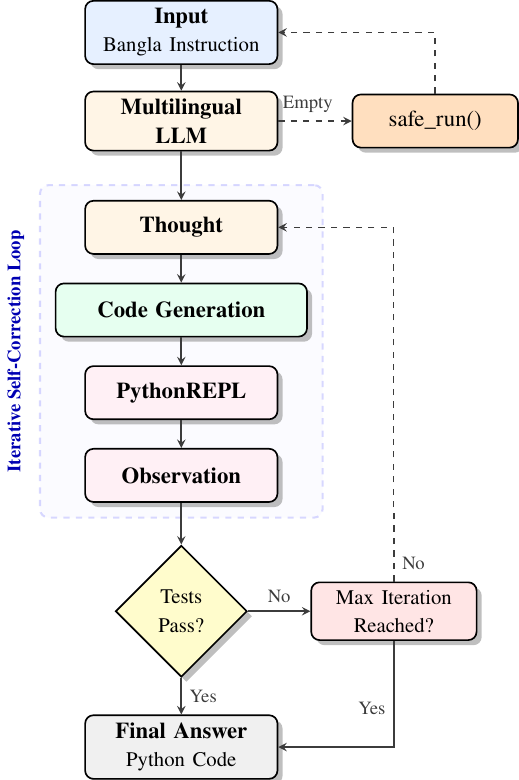}
    \caption{Thought-Code-Observation Cycle in the BanglaCodeAct Agent Framework. This diagram illustrates the iterative process of generating code, executing it, providing feedback, and refining the solution based on self-correction, facilitating cross-lingual code generation in Bangla.}
    \label{fig:codeact-methodology}
\end{figure}





\begin{enumerate}
    \item \textbf{Thought:} The agent generates an internal monologue, outlining its understanding and plan for the task in \texttt{<thought>}, showcasing reasoning in Bangla before code generation.
    
    \item \textbf{Code Generation:} The agent produces Python code based on the plan, enclosed in \texttt{<code>} with test assertions for immediate self-verification.

    \item \textbf{Execution and Feedback:} The code runs in a sandboxed \textbf{PythonREPL} with a timeout. Errors, like \texttt{TypeError}, provide feedback for \textbf{iterative self-correction}, refining the solution until valid or a max iteration is reached, with the result in \texttt{<answer>}.
\end{enumerate}

To enhance reliability, we implement a \textbf{retry handler (\texttt{safe\_run})} that re-initiates the reasoning process if the agent produces an invalid or empty response. The retry mechanism permits a user-defined number of task attempts, improving success rates by reducing sporadic failures.

\begin{table}[!ht]
    \centering
    \scriptsize
    \begin{tabular}{@{}p{0.18\textwidth} p{0.24\textwidth}@{}}
        \toprule
        \textbf{Instruction (Bengali)} & \textbf{Test Cases} \\ \midrule
        {\bng is/TRNNG ethek pRdt/t AkKerr pRthm EbNNG eshSh Upsr/g muech ephlun.} 
        Example: \texttt{remove\_Occ(s, ch)} &
        \begin{tabular}[t]{@{}l@{}}
            remove\_Occ("hello","l") == "heo" \\
            remove\_Occ("banana","a") \newline == "bann" \\
            remove\_Occ("abc","x") == "abc"
        \end{tabular} \\
        \midrule
        {\bng EkiT pRdt/t mYaiTRk/sek tar sairguilr eJagphl AnuJaJii sajan.}
        Example: \texttt{sort\_matrix(M)} &
        \begin{tabular}[t]{@{}l@{}}
            sort\_matrix([[1,2,3],[2,4,5],[0,1,1]]) \\
            \hspace{5mm} == [[0,1,1],[1,2,3],[2,4,5]] \\
            sort\_matrix([[5,5],[2,2],[3,3]]) \\
            \hspace{5mm} == [[2,2],[3,3],[5,5]]
        \end{tabular} \\
        \bottomrule
    \end{tabular}
    \caption{Illustrating error recovery in ambiguous and complex cases.}
    \label{tab:qualitative-example}
\end{table}

For instance, {\bng ``is/TRNNG ethek pRdt/t AkKerr pRthm EbNNG eshSh Upsr/g muech ephlun''} (remove the first and last occurrence of a given character from a string). Initial attempts produced incomplete logic (removing only one occurrence). (see Table~\ref{tab:qualitative-example}).

\section{Results and Analysis}
\label{sec:results}

Different models and experiments were conducted during the development phase, which are reported in ~\ref{appendix:Baselines}. The experiment setup and hyperparameter details are described in table ~\ref{tab:hyper}.

The `pass@1` scores for all evaluated methods on the mHumanEval dataset are summarized in Table \ref{tab:results}. Our proposed BanglaCodeAct Agent achieves a `pass@1' score of \textbf{94.0\%}, significantly outperforming all other methods.

\begin{table}[h!]
    \centering
    \small
    \begin{tabular}{@{}l l c@{}}
        \toprule
        \rowcolor[HTML]{CFE2F3} 
        \textbf{LLM Model} & \textbf{Method} & \textbf{pass@1} \\ \midrule
        \rowcolor[HTML]{F9F9F9} 
        Qwen3-8B & \textbf{BanglaCodeAct} & \textbf{\cellcolor[HTML]{D9EAD3} 94.0} \\
        Qwen3-8B & Self-Consistency & 88.0 \\
        Qwen3-8B & Majority Voting & 66.0 \\
        Qwen3-8B & Few-Shot & 46.0 \\
        Qwen3-8B & Zero-Shot & 36.0 \\
        \rowcolor[HTML]{F9F9F9} 
        Qwen2.5-Coder-7B & Few-Shot & 51.0 \\
        Qwen2.5-Coder-7B & Zero-Shot & 44.0 \\
        Llama-3.1-8B & Zero-Shot & 39.0 \\
        Llama-3.1-8B & Few-Shot & 77.0 \\
        \rowcolor[HTML]{F9F9F9} 
        DeepSeek-Coder-V2-Lite & \textbf{BanglaCodeAct} & \textbf{\cellcolor[HTML]{D9EAD3} 73.8} \\
        DeepSeek-Coder-V2-Lite & Few-Shot & 73.0 \\
        DeepSeek-Coder-V2-Lite & Zero-Shot & 71.4 \\
        TigerLLM-1B-it & Zero-Shot & 11.0 \\
        \bottomrule
    \end{tabular}
    \caption{Comparison of pass@1 accuracy (\%) for different models and prompting strategies on the mHumanEval dataset. Our proposed \textbf{BanglaCodeAct Agent} (Qwen3-8B) achieves the highest score, demonstrating the effectiveness of iterative self-correction.}
    \label{tab:results}
\end{table}


The results in Table \ref{tab:results}, clearly demonstrate the efficacy of our agent-based framework.
\newline




The experimental results demonstrate the effectiveness of the proposed \textbf{BanglaCodeAct Agent} in leveraging an iterative self-correction mechanism for Bangla-to-Python code generation. With the Qwen3-8B model, the agent achieves a \textbf{94.0\% pass@1 accuracy}, significantly outperforming Zero-Shot (36.0\%), Few-Shot (46.0\%), and Majority Voting (66.0\%) strategies (Table~\ref{tab:results}). It ranked 17th on the test set (71.6\%) and 8th on the development set (94\%).

These results underscore the agent’s ability to correct common code generation errors using REPL feedback, which distinguishes it from static prompting approaches. The Qwen3-8B model outperforms specialized models like Qwen2.5-Coder-7B, highlighting the importance of multilingual reasoning over code-specific training. Primary failure cases occur with semantically ambiguous or complex instructions, where the agent may not converge within 10 iterations.

\section{Limitations}

Despite strong performance, BanglaCodeAct has several limitations. The model's effectiveness is limited by its size, and experiments with larger LLMs (e.g., 32B parameters or more) were not conducted due to GPU resource constraints. Such models could potentially improve code generation accuracy for more complex tasks.

Additionally, the current evaluation primarily focuses on algorithmic and syntactic correctness. The system's ability to understand semantics and handle ambiguous or context-dependent Bangla instructions remains an open challenge. Moreover, the system relies on high-quality test cases for feedback, which may not always be available in real-world scenarios. The performance could be further limited by the absence of such reliable test cases in practice.

\bibliography{references.bib}

@inproceedings{bhattacharjee2023banglabert,
  title={{BanglaBERT: A Denoising Autoencoding based Pre-trained Language Model for Bengali}},
  author={Bhattacharjee, Abhishek and Aktar, Shammur and Zishan, Md Saidul Islam and Ghosh, Sudipta and Hasan, Md Mahadi and Rahman, M Sohel and Alam, Mohammad Ruhul and Nakagawa, Tetsuji and Misu, Teruhisa and Shah, Farig},
  booktitle={Proceedings of the 17th Conference of the European Chapter of the Association for Computational Linguistics},
  pages={2577--2591},
  year={2023}
}

@article{chen2021evaluating,
  title={{Evaluating Large Language Models Trained on Code}},
  author={Chen, Mark and Tworek, Jerry and Jun, Heewoo and Yuan, Qiming and de Oliveira Pinto, Henrique P. and Kaplan, Jared and Edwards, Harri and Burda, Yuri and Joseph, Nicholas and Brockman, Greg and others},
  journal={arXiv preprint arXiv:2107.03374},
  year={2021}
}

@article{guo2024deepseek,
  title={DeepSeek-Coder: When the Large Language Model Meets Programming--The Rise of Code Intelligence},
  author={Guo, Daya and Zhu, Qihao and Yang, Dejian and Xie, Zhenda and Dong, Kai and Zhang, Wentao and Chen, Guanting and Bi, Xiao and Wu, Yu and Li, YK and others},
  journal={CoRR},
  year={2024}
}

@inproceedings{kwon2023efficient,
  title={Efficient memory management for large language model serving with pagedattention},
  author={Kwon, Woosuk and Li, Zhuohan and Zhuang, Siyuan and Sheng, Ying and Zheng, Lianmin and Yu, Cody Hao and Gonzalez, Joseph and Zhang, Hao and Stoica, Ion},
  booktitle={Proceedings of the 29th Symposium on Operating Systems Principles},
  pages={611--626},
  year={2023}
}

@article{lewisretrieval,
  title={Retrieval-Augmented Generation for Knowledge-Intensive NLP Tasks},
  author={Lewis, Patrick and Perez, Ethan and Piktus, Aleksandra and Petroni, Fabio and Karpukhin, Vladimir and Goyal, Naman and K{\"u}ttler, Heinrich and Lewis, Mike and Yih, Wen-tau and Rockt{\"a}schel, Tim and others},
  journal={arXiv preprint arXiv:2005.11401},
  year={2020}
}

@article{li2022alphacode,
  title={{Competition-level Code Generation with AlphaCode}},
  author={Li, Yujia and Choi, David and Chung, Junyoung and Kushman, Nate and Schrittwieser, Julian and Dadashi, Rémi and Lazic, Dora and Veli{\v{c}}kovi{\'c}, Petar and Son, Jiayuan and Botha, Johanni and others},
  journal={Science},
  volume={378},
  number={6624},
  pages={1092--1097},
  year={2022},
  publisher={American Association for the Advancement of Science}
}

@article{li2023starcoder,
  title={{StarCoder: May the Source be With You!}},
  author={Li, Raymond and Allal, Léonard Blanchard and Zi, Yuchen and Muennighoff, Niklas and Kocetkov, Denys and Mou, Chenglei and Marone, Marc and Akiki, Christian and Li, Jian and Chim, Jenny and others},
  journal={Transactions on Machine Learning Research},
  year={2023},
  publisher={OpenReview}
}

@article{raihan2025tigercoder,
  title={TigerCoder: A Novel Suite of LLMs for Code Generation in Bangla},
  author={Raihan, Nishat and Anastasopoulos, Antonios and Zampieri, Marcos},
  journal={arXiv preprint arXiv:2509.09101},
  year={2025}
}

@inproceedings{raihan-etal-2025-blp,
  title={Overview of {BLP}-2025 Task 2: Code Generation in Bangla},
  author={Raihan, Nishat and Jawad, Mohammad Anas and Rahman, Md Mezbaur and Ulfat, Noshin and Gupta, Pranav and Rahman, Mehrab Mustafy and Karmakar, Shubhra Kanti and Zampieri, Marcos},
  booktitle={Proceedings of the Second Workshop on Bangla Language Processing (BLP-2025)},
  month=dec,
  year={2025},
  publisher={Association for Computational Linguistics}
}

@inproceedings{raihan-etal-2025-mhumaneval,
  title={mHumanEval: A Multilingual Benchmark to Evaluate Large Language Models for Code Generation},
  author={Raihan, Nishat and Anastasopoulos, Antonios and Zampieri, Marcos},
  editor={Chiruzzo, Luis and Ritter, Alan and Wang, Lu},
  booktitle={Proceedings of the 2025 NAACL Conference (Long Papers)},
  month=apr,
  year={2025},
  address={Albuquerque, New Mexico},
  publisher={Association for Computational Linguistics},
  pages={11432--11461},
  doi={10.18653/v1/2025.naacl-long.570},
  url={https://aclanthology.org/2025.naacl-long.570/},
  ISBN={979-8-89176-189-6}
}

@article{roziere2023code,
  title={{Code Llama: Open Foundation Models for Code}},
  author={Roziere, Baptiste and Gehring, Jonas and Gloeckle, Fabian and Tworkowski, Sten and Lachaux, Marie-Anne and Lavril, Thibaut and Mason, Iz and Masson, Alexandre and Mensch, Arthur and others},
  journal={arXiv preprint arXiv:2308.12950},
  year={2023}
}

@article{vaswani2017attention,
  title={Attention is all you need},
  author={Vaswani, Ashish and Shazeer, Noam and Parmar, Niki and Uszkoreit, Jakob and Jones, Llion and Gomez, Aidan N and Kaiser, Lukasz and Polosukhin, Illia},
  journal={Advances in Neural Information Processing Systems},
  volume={30},
  year={2017}
}

@article{yan2024corrective,
  title={Corrective Retrieval-Augmented Generation},
  author={Yan, Shi-Qi and Gu, Jia-Chen and Zhu, Yun and Ling, Zhen-Hua},
  journal={Available at SSRN 5267341},
  year={2024}
}

@article{yin2022natural,
  title={Natural language to code generation in interactive data science notebooks},
  author={Yin, Pengcheng and Li, Wen-Ding and Xiao, Kefan and Rao, Abhishek and Wen, Yeming and Shi, Kensen and Howland, Joshua and Bailey, Paige and Catasta, Michele and Michalewski, Henryk and others},
  journal={arXiv preprint arXiv:2212.09248},
  year={2022}
}

@article{zhong2024opportunities,
  title={Opportunities and challenges of large language models for low-resource languages in humanities research},
  author={Zhong, Tianyang and Yang, Zhenyuan and Liu, Zhengliang and Zhang, Ruidong and Liu, Yiheng and Sun, Haiyang and Pan, Yi and Li, Yiwei and Zhou, Yifan and Jiang, Hanqi and others},
  journal={arXiv preprint arXiv:2412.04497},
  year={2024}
}

\end{document}